\relax
\documentclass[runningheads,a4paper]{llncs}
\usepackage{times}  
\usepackage{helvet}  
\usepackage{courier}  
\usepackage{url}  
\usepackage{graphicx}  
\usepackage{dcolumn}

\usepackage{comment}
\usepackage{mathtools}
\usepackage{tabularx, multirow}
\usepackage{colortbl, color}
\usepackage{xcolor}
\usepackage{booktabs}
\usepackage{tikz}
\usepackage{dblfloatfix}
\usepackage{multicol}
\usepackage{subfig}
\usepackage{ccicons}

\title{TokTrack: A Complete Token Provenance and Change Tracking Dataset for the English Wikipedia}

\author{Fabian Fl\"{o}ck \inst{1} \and Kenan Erdogan \inst{1} \and Maribel Acosta \inst{2}}
  \authorrunning{Fabian Fl\"{o}ck et al.}
  \titlerunning{TokTrack: A Complete Token Provenance and Change Tracking Dataset}
 \institute{GESIS - Leibniz Institute for the Social Sciences, Germany \and Karlsruhe Institute of Technology
}

\begin{document}

\maketitle

\begin{abstract}
We present a dataset that contains every instance of all tokens ($\approx$ words) ever written in undeleted, non-redirect English Wikipedia articles until October 2016, in total $13,545,349,787$ 
instances. Each token is annotated with (i) the article revision it was originally created in, and (ii) lists with all the revisions in which the token was ever deleted and (potentially) re-added and re-deleted from its article, enabling a complete and straightforward tracking of its history.
This data would be exceedingly hard to create by an average potential user as it is (i) very expensive to compute and as (ii) accurately tracking the history of each token in revisioned documents is a non-trivial task. 
Adapting a state-of-the-art algorithm, we have produced a dataset that allows for a range of analyses and metrics, already popular in research and going beyond, to be generated on complete-Wikipedia scale; ensuring quality and allowing researchers to forego expensive text-comparison computation, which so far has hindered scalable usage.
We show how this data enables, on token-level, computation of provenance, measuring survival of content over time, very detailed conflict metrics, and fine-grained interactions of editors like partial reverts, re-additions and other metrics, in the process gaining several novel insights. 

\end{abstract}


\section{Introduction}
In collaborative writing platforms like Wikipedia, every single revision is recorded and, at least for the Wikimedia projects, made publicly available.
Research has enthusiastically made use of these fine-grained edit logs  
to study cooperation, conflict, and other phenomena (see Section \ref{sec:relwork} for related work). 
The same data has also been investigated regarding the characteristics of the content itself, e.g. in terms of what makes textual content survive in a Wiki or what content becomes the subject of controversies. These questions are all connected to the collaboration dynamics that give rise to the content.

The majority of the studies in these areas have concentrated on the article pages of the English language edition of Wikipedia as the use case for collaborative writing platforms, and so
our dataset is built on the English Wikipedia article data as well.
\footnote{``(Wikipedia) articles'' are all pages in the main namespace of Wikipedia (ns=0), in contrast to, e.g., article talk pages, user (talk) pages, pages about guidelines and rules, etc.}
In particular, the research we discuss here has been conducted using the complete history of edits applied to an article over time. An edit creates a new revision (or version) of the article, which is saved in its entirety in Wikipedia's database (instead of the differences between revisions). 
Some of the information needed to conduct such studies is relatively easy to access computationally from the XML database dumps provided by the Wikimedia foundation -- and potentially any standard installation of Mediawiki.\footnote{Mediawiki is the most widely used Wiki software, also deployed for most 
of the Wikimedia projects.} Often used are, e.g., the sequence of edits plus their metadata such as timestamps, editor or byte length, and the identification of revisions that reset content to a revision with identical content to compute reverts.

\begin{figure}[t!]
\centering
\includegraphics[width=0.93\linewidth]{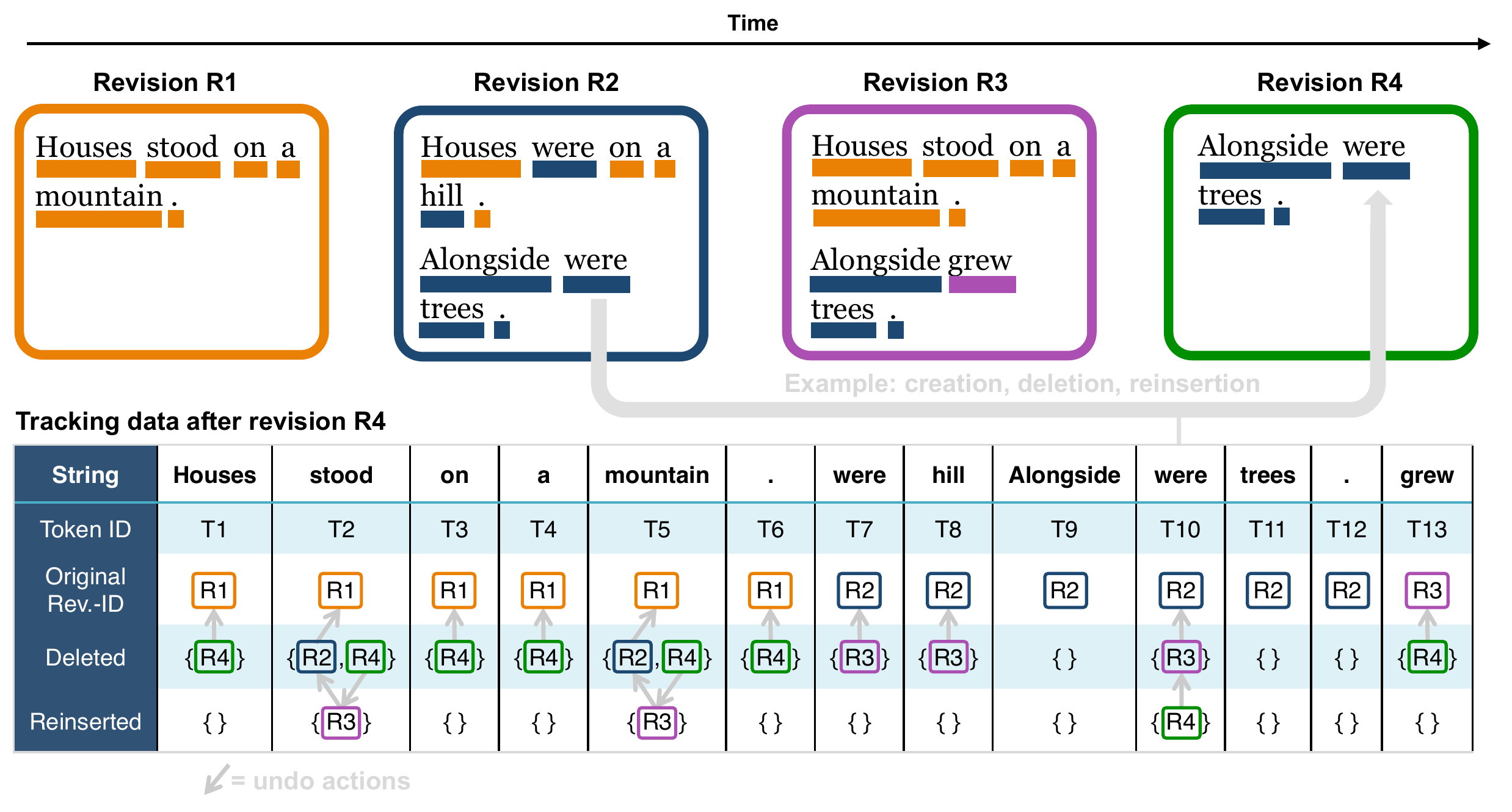}

\caption{\textbf{Toy example for token tracking.} Top: Sequential revisions R1, R2, R3, and R4 of a dummy article from left to right. Color of underlining indicates origin revision. Bottom: The token tracking data after revision R4, showing all tokens that have so far appeared in the article, if currently deleted or not. Each token is linked to a unique token ID (note some identical string values for distinct tokens), the revision ID of original introduction, and each revision in which a token was possibly a) deleted and b) reinserted (both are lists). Arrows in the table indicate undo actions among revisions.}
\label{fig:trackxample}
\end{figure}

Yet, more in-depth data that ``zooms in'' on the exact changes in each revision are more expensive to extract, but can highly benefit research, as we will outline further in Section \ref{sec:relwork}. The caveat is that such analysis needs to process alterations  applied at the level of individual tokens, usually defined as strings separated by white-spaces, i.e., what is commonly called ``words". Additionally, special characters (``\%", ``.", etc.) can also be extracted as tokens and are used to delimit words.  
However, to detect the changes to these tokens and their original provenance in the revision history necessitates the application of computationally expensive text-comparison algorithms, which on top can often be inaccurate when tracking alterations to all tokens in 
each edit \cite{deAlfaro:2013}. 
Unfortunately, as of today, no publicly available dataset exists that would eliminate the need for 
computing the changes of tokens at large scale while guaranteeing high accuracy, 
which is likely one reason why many studies dealing with token-level changes do not analyze the full revision history dump, but rather constrain themselves to subsets of articles (cf. Section \ref{sec:relwork}). 

Consequently, we created TokTrack, a dataset that tracks the origin and changes of all tokens in the articles of the English Wikipedia. In the remainder we first outline the process to create TokTrack, its structure and how it can be retrieved (Section \ref{sec:dataset}). 
Thereafter, we summarize research strands that might profit from using our dataset in the future  (Section \ref{sec:relwork}). 
Then, we conduct some descriptive examinations that showcase the possibilities for analyses of the dataset that formerly would have been much more complicated or simply unfeasible to achieve (Section \ref{sec:usecases}), and provide some auxiliary datasets derived in the process. Our findings also reveal some new insights
regarding authorship, content survival, reverts, and conflict on Wikipedia.

\section{The TokTrack Dataset}
\label{sec:dataset}
\subsection{Origin}
The TokTrack dataset was created from the full English Wikipedia revision history data dump of November 1, 2016, as published by Wikimedia, and hence includes all complete months from January 2001 to October 2016.\footnote{\url{https://dumps.wikimedia.org/enwiki/20161101/}} 
This XML dump contains the full content (in Wiki Markup) of revisions for all pages that were not deleted.\footnote{I.e., content that was completely removed and is not accessible anymore in Wikipedia.} 
Uncompressed, the XML dump measures around 12 TB. 
We extracted the article pages (namespace $=0$) 
that were not redirect pages\footnote{\url{https://en.wikipedia.org/wiki/Wikipedia:Redirect}}  as per their latest revision contained in the XML dump. 
This resulted in $5,275,388$ articles being extracted. 
For each article, we obtained the corresponding revisions. In total, 451,350,901 
revisions are contained in the TokTrack dataset. 
  
\subsection{Dataset Creation}
\label{sec:datacreation}

\subsubsection{The challenge of tracking tokens.}
A token constitutes the atomic unit of the TokTrack dataset and can, as mentioned,  consist of words or special characters. 
A token is a specific instance of an immutable string, e.g., ``house", while another token in the same or a different article can share the same string value. 
A token is unique in an article and can, after it is originally added in one revision, appear in several revisions, even when disappearing (after being deleted) for some revisions before re-appearing again (being reinserted). 

To illustrate, consider the example in Figure \ref{fig:trackxample}, and there specifically the case of token with ID T10: This token (string value ``were") should be tracked as being first added in revision R2, deleted in revision R3 and reinserted in revision R4, instead of treating it as a new token in revision R4.  
This poses the first challenge, as solely applying a text comparison algorithm from each revision to the previous one cannot detect such reinsertions \cite{javanmardi2010modeling}. 
The second challenge is that in a real Wikipedia article, a sequence of tokens can be moved to a ``remote'' position of the article, in which most text-difference algorithms will lose track and treat the tokens as deleted and ``new'' tokens with the same string values as being introduced.

\subsubsection{Available solutions.}
Several approaches exist that are in principle suited to tackle the described challenges. 

One technique applied by some works to compute provenance and survival of tokens over revisions  is to, on one hand, compute the changes from one revision to the immediately previous one using an out-of-the-box text-difference algorithm \cite{priedhorsky_creating_2007,halfaker_jury_2009}. 
Secondly, to check for deleted words added back by an editor who not originally wrote them, the approach also checks for so-called ``identity reverts", which describe the reset of the article to an exactly identical version of the content. 
If such a revert occurs, the provenance of the tokens is attributed to the original revisions and authors, not the restorer. 
These identity reverts seem to be the most common type of reverts \cite{kittur_he_2007}. 
However, as Suh et al. \cite{suh2007usvsthem} mention: ``the disadvantage of this method is that it does not pick up partial reverts, in which only some of the text in an article is reverted'' and an identical revision is not created; see for example Figure \ref{fig:trackxample}, where we did not include any revision's content reoccurring identically in another revision. 
Likewise, 
 Brandes et al. \cite{brandes2} remark that using only identity reverts does lead to inaccuracies. 
 We will see in Section \ref{sec:usecases} how many partial reverts actually occur. 

The \emph{Wikitrust} approach by Adler et al. \cite{adler_content-driven_2007,Adler_measuring_2008} detects provenance by searching for longest matches for all word sequences of the current revision in selected preceding revisions and their previously existing (but now deleted) word-chunks. 
As De Alfaro and Shavlovsky \cite{deAlfaro:2013} later argue, it is not well-suited for the task of authorship or provenance detection, as the process depends on several factors of its ``computationally involved" editor reputation calculation -- a suspicion supported by an evaluation on a small sample of authorship data generated with \textit{Wikitrust}, yielding only  around 50\% correctly attributed authors for tokens \cite{floeck_whose_2012}.  

De Alfaro and Shavlovsky \cite{deAlfaro:2013} in their recent work propose a more advanced technique for attributing original authorship to the tokens in a target revision of an article which can be also used to infer changes applied to a token over its lifetime. 
This approach exploits a ``trie" structure to check for recurring $n$-gram token sequences of the currently analyzed revision in the content of all previous revisions. 
It therefore does not rely on identical revisions to be existent and can detect even partial reinsertions. 
It was shown that this approach scales well even to very large articles. 


More recently, we have proposed another approach, \textit{Wikiwho}, aimed at calculating token provenance, and also able to perform change detection \cite{wikiwho}. \textit{Wikiwho} splits revisions into paragraphs and sentences and then checks for identical reuse of these smaller article parts, which is combined with a standard text-diff in cases where no matches are found. 
In comparison with the approach by \cite{deAlfaro:2013}, \textit{Wikiwho} achieves distinctly better runtimes \cite{wikiwho}. 
In regard to precision of attributing original contributors to tokens, experimental results indicate that \textit{Wikiwho} outperforms the algorithm by \cite{deAlfaro:2013} and reaches 95\% correct attributions. 
To the best of our knowledge, our prior work was also the only one so far to  evaluate related techniques regarding the correctness of their results. 

\subsubsection{Data processing.} 
We chose \textit{Wikiwho} to process the XML dump and extended it to additionally produce the \texttt{token ID}, \texttt{deleted} and \texttt{reinserted} data per token as shown in the lower part of Figure \ref{fig:trackxample}. 
Each article's revision history is processed individually.\footnote{Tracking content moves across articles is not supported out-of-the-box in any solution and would also require exponentially more computational resources to achieve.}  
The algorithm parses the sequence of the revisions ordered by timestamp, comparing the content of each revision with all its predecessors.  
Tokens include all wiki markup, e.g., ``[" and ``]" (used in link syntax), while string capitalization is ignored. 
The article content has been tokenized as per the original Wikiwho settings including some refinements; white spaces and new lines are used as delimiters (but not considered tokens), while special characters are both used as delimiters and tokens.

The processing of all English Wikipedia articles (including appropriate storage in a data base for analysis) was completed after around $25$ days on a dedicated Ubuntu Server 16.04 VM with 122 GB RAM and 20 cores. 



\subsection{Dataset Size and Structure}

In total, out of the $5,275,388$ articles processed, $13,545,349,787$ 
instances of tokens were extracted. 
These include the tokens still present in articles in the last revision of the Wikipedia XML dumps, i.e., the most current one as of processing (analogous to tokens T9-T12 in Figure \ref{fig:trackxample}), as well as all tokens that were ever added, but subsequently deleted from the articles (analogous to tokens with T1-T8 and T13 in Figure \ref{fig:trackxample}).

Given this particular nature of the data, we have generated three different \emph{output types} per article, structured as follows:
\begin{itemize}
	\item \textbf{current\_content}:  Contains the tokenized content of all articles present in the last revision of the XML dump, as of Nov. 1, 2016.  
	Each line contains one token with the following fields, analogous to our example in Figure \ref{fig:trackxample}: 
\begin{itemize}
\item \texttt{page\char`_id}  (integer scalar): The page ID of the article (as extracted from the XML dumps) to which the token belongs.
\item \texttt{last\char`_rev\char`_id} (integer scalar): The revision ID  where the token \textit{last appeared}; in this output type, this is the last revision ID included in the downloaded XML dumps as of November 1, 2016 per each article.
\item \texttt{token\char`_id} (integer scalar): The token ID assigned internally by the algorithm, unique per article. Token IDs are assigned increasing from 1 for each new token added to an article. 
\item \texttt{str} (string value): The string value of the token.
\item \texttt{origin\char`_rev\char`_id} (integer scalar): The ID of the revision where the token was added originally in the article. 
\item  \texttt{out} (ordered integer list): List of all revisions in which the token was \textit{deleted}, ordered sequentially by time.

\item  \texttt{in} (ordered integer list): List of all revisions where the token was \textit{reinserted} after being deleted previously, ordered sequentially by time. One \texttt{in} has to be preceded by one \texttt{out} in sequence. E.g. in Fig. \ref{fig:trackxample}, token T10 was created in R2, deleted in R3, and reintroduced in R4.
%

%

\end{itemize}

\item \textbf{deleted\_content}: Contains all tokens that have ever been present in articles in at least one revision, but were not present anymore for the last revision in the XML dump. The structure of the file is exactly equivalent to \textbf{current\_content} with two differences: (1) At least one entry exists in the \texttt{out} list of each token and one more \texttt{out} than \texttt{in}. (2) The \texttt{last\char`_rev\char`_id} field can contain different values for  tokens of the \textit{same} article, as deleted tokens might have appeared last at different revisions.
\item  \textbf{revisions}: Lists all revisions of the articles as processed by the algorithm in sequential order. The contained information can be joined with the other two file types on the \texttt{origin\char`_rev\char`_id} or \texttt{last\char`_rev\char`_id} fields. Each line represents one revision, including metadata: 
\begin{itemize}
    \item \texttt{page\char`_id}  (integer scalar): The page ID of the article (as extracted from the XML dumps) to which the revision belongs.
	\item \texttt{rev\char`_id} (integer scalar): The revision ID. Revision IDs are extracted from the XML dumps, belong to one article only and are unique for the whole dataset. 
	\item   \texttt{timestamp} (timestamp): The creation timestamp of the revision as extracted from the XML dumps.
	\item \texttt{editor} (string value): The user ID of the editor as extracted from the XML dumps.
 	User IDs are integers, are unique for the whole Wikipedia and can be used to fetch the current name of a user. The only exemption is user ID $=0$, which identifies all unregistered accounts. To still allow for distinction between unregistered users, the identifiers of unregistered users are included in this field, prefixed by ``0$\vert$".

	\end{itemize}
\end{itemize}

We provide these files in CSV format, with each CSV file being a batch containing a certain range of articles by page id. File names have the structure: 
\texttt{<XML dump date>\char`-<output type>\char`-<batch id>\char`-<first page id in batch>\char`-\\<last page id in batch>.csv} 
These files CSV files are again bundled in compressed archives, with names indicating contained batches. 
The total size of the dataset is 763 GB uncompressed and 69 GB compressed. 
The TokTrack dataset is available for download 
\cite{floeck2017toktrack_data}.\footnote{Direct link: \url{https://zenodo.org/record/345571}} 
We have also stored the results in a Postgres database and can produce different outputs on request.

The reason for splitting ``current\_content" and ``deleted\_content" output types is that we expect many use cases to involve analyses of the tokens of the most current revision where access to deleted tokens is not required. 
If needed, however, the two output types can easily be used in combination. In principle, it is possible to recreate the exact content present in each revision of every article trough the information stored in the \texttt{out} and \texttt{in} lists and compute metrics like, e.g., conflict for that specific revision. Through the \texttt{rev\char`_id} and \texttt{page\char`_id} fields, further metadata can be retrieved if needed via the Wikipedia dumps or API.\footnote{\url{https://www.mediawiki.org/wiki/API:Main_page}}

Alongside this core dataset, we also publish auxiliary datasets computed on top of it, which are used in the analyses in Section \ref{sec:usecases}.

As an additional service, we will also make parts of the data covered here accessible through a Web API, based on the latest article content of the English Wikipedia.\footnote{\url{http://api.wikiwho.net/api/}}

\section{Research Using Data Derivable from the TokTrack Dataset} 
\label{sec:relwork}

A range of research strands exist that either rely on data that can be derived from the TokTrack dataset in a much more efficient manner or could profit otherwise from it. 
We also argue that a canonical content-tracking dataset for Wikipedia, produced with a quality-tested algorithm would enable the execution and reproducibility of many future studies.
In the following we analyze publications that address 
related topics by the kind of data they have been extracting and using to accomplish their objectives, and discuss what benefits future investigations in a similar vein could gain from our dataset.




\subsection{Authorship, Persistence, and  Reputation} 

The first group of approaches can be roughly identified by their shared aim to detect (i) the revision of original introduction of a token (and by extension, the original author, timestamp, etc.) and (ii) the subsequent survival or persistence of that token in the article, either measured in time or number of revisions. 
This requires to keep track of potential deletions and re-additions of a token after the initial addition, as the same tokens can be deleted and reinserted multiple times over the course of the article writing process, while still being originally written by the same author. 

Priedhorsky et al. \cite{priedhorsky_creating_2007} 
use a \textit{content persistence} metric to identify which editors ``owned'' how much content in every revision and to determine whether an editor's content gets a certain amount of views, while Halfaker et al. \cite{halfaker_jury_2009,halfaker2011dont} 
approximate the quality of an edit and the productivity of editors with an equivalent approach. 
These techniques rely on a combination of identity reverts and a text-difference algorithm to track persistence (cf. Section \ref{sec:datacreation}).
In order to compute the reputation of editors as well as the trustworthiness of their content, the \emph{Wikitrust} approach by Adler et al. \cite{adler_content-driven_2007,Adler_measuring_2008} detects provenance and persistence and also features an interface for coloring more or less trusted content sequences. 
Javanmardi et al. \cite{javanmardi2010modeling} follow a similar strategy for computing user reputation, but use only identity reverts aside from text-difference computation.
The \emph{HistoryFlow} visualization \cite{viegas_2004} also makes use of tracking the authors and positions in each revision of sample articles, although on the (coarse-grained) sentence-level, drawing them in a so-called stratigraph. The technique provides a visual ``story'' of an article's writing history and has been reproduced in several community projects.\footnote{Cf., e.g., \url{http://fogonwater.com/blog/2015/11/wikipedia-edit-history-stratigraphy} or \url{http://iphylo.blogspot.de/2009/09/visualising-edit-history-of-wikipedia.html}} \  

From the description in Section \ref{sec:dataset} it becomes clear that similar future analyses and tools could skip the tedious and expensive process of precomputing the needed data, be it with self-built or reused text comparison approaches, by simply extracting provenance and calculating survival from the explicit markers in our dataset.

\subsection{Identifying Conflicted Content}

A recurring theme in Wikipedia-related research is the measurement and characterization of the conflict or controversy specific content is subject to.

For one part, studies have focused on the particular problem of identifying controversial articles as a whole. Either by computing the mutual (identity) revert levels between editors \cite{sumi2011edit,yasseri2012dynamics,yasseri2014most}; 
or by building more complex models, involving a range  of metrics beside reverts, e.g., talk page and anonymous edits, edit comments and removed words \cite{kittur_he_2007,vuong_ranking_2008,sepehri2012leveraging}. 
More recently, researchers have also been concerned with pinpointing the specific parts of an article that are subject to controversy between editors. Borra et al. \cite{contropedia2} developed the web platform \textit{Contropedia} to visualize controversies related to the internal Wikipedia-links present in an article, by coloring them in different shades representing their conflict levels. 
Bykau et al. \cite{bykaufine} proposed a novel method to cluster disagreements of editors over specific tokens into larger controversies, delimited not only by the content involved, but also determining a timeframe for when a controversy occurs. 

The TokTrack dataset can especially be of use for the second type of  research, which relies on determining how often a token was deleted, reinserted, deleted again and so on. 
Our dataset allows to extract this information easily from the \texttt{out} and \texttt{in} lists of each token. Further information about, e.g., time-differences (to detect rapid interchanges) or editors (to detect mutual or self-targeted delete/reinsert actions) can be retrieved from the \textit{revisions} file. 
The materialized computations of token changes in TokTrack enable research to detect conflicted content sequences at large scale (the current approaches do not use large samples). 

On the other hand, the methods to identify controversial articles seem to already enable reliable analyses of large quantities of documents by drawing on identity reverts and other features that are more straightforward to extract than individual token histories. However, we hope that the TokTrack dataset can pave the road towards refinements of established approaches, which may be able to take advantage of the more detailed change data we make available.

\subsection{Interactions Between Editors}

A wide range of studies have used the edit history of Wikipedia articles to draw conclusions about the interactions between editors through their actions applied on content, in order to learn about the collaborative process. The first type of commonly used input data is  the time-ordered \textbf{sequence of edits} (and their associated metadata), where edits following another editor's actions  under certain conditions are translated into specific interactions (e.g.,  \cite{iba2010analyzing,keegan2016analyzing}).
 Extracting these sequences from the Wikipedia dumps is relatively straightforward, so that such analyses do not benefit directly from the data we provide.
 
A second type of input are \textbf{identity reverts} of the article to a duplicate former content state, which are interpreted as an antagonistic action by the reverting user towards the originators of the edits between the two identical revisions. 
Some works that use identity reverts as a central tool provide a general overview of the evolution of editing dynamics and the changing state of Wikipedia's editor community 
\cite{kittur_he_2007,suh_singularity_2009}.	
Other works focus on reverts in regard to their indicator role of a reverted edit's (insufficient) quality \cite{halfaker_jury_2009} as well as their possible damaging effects and the harmful barriers faced by new editors  \cite{halfaker2011dont,halfaker2012rise}. Gender imbalances in how editor's work is received \cite{clubhouse} have been investigated with the help of reverts
 as well as 
 more intricate motifs of editor interaction  \cite{dedeo2015conflict,tsvetkova2016dynamics}. 

\begin{figure}[t!]

\subfloat[Tokens newly added per month, split into tokens (i) that didn't survive 48 hours, (ii) that survived 48h, but not until (end of) Oct.'16 and (iii) that survived until Oct.'16. ]{
\includegraphics[width=0.29\linewidth]{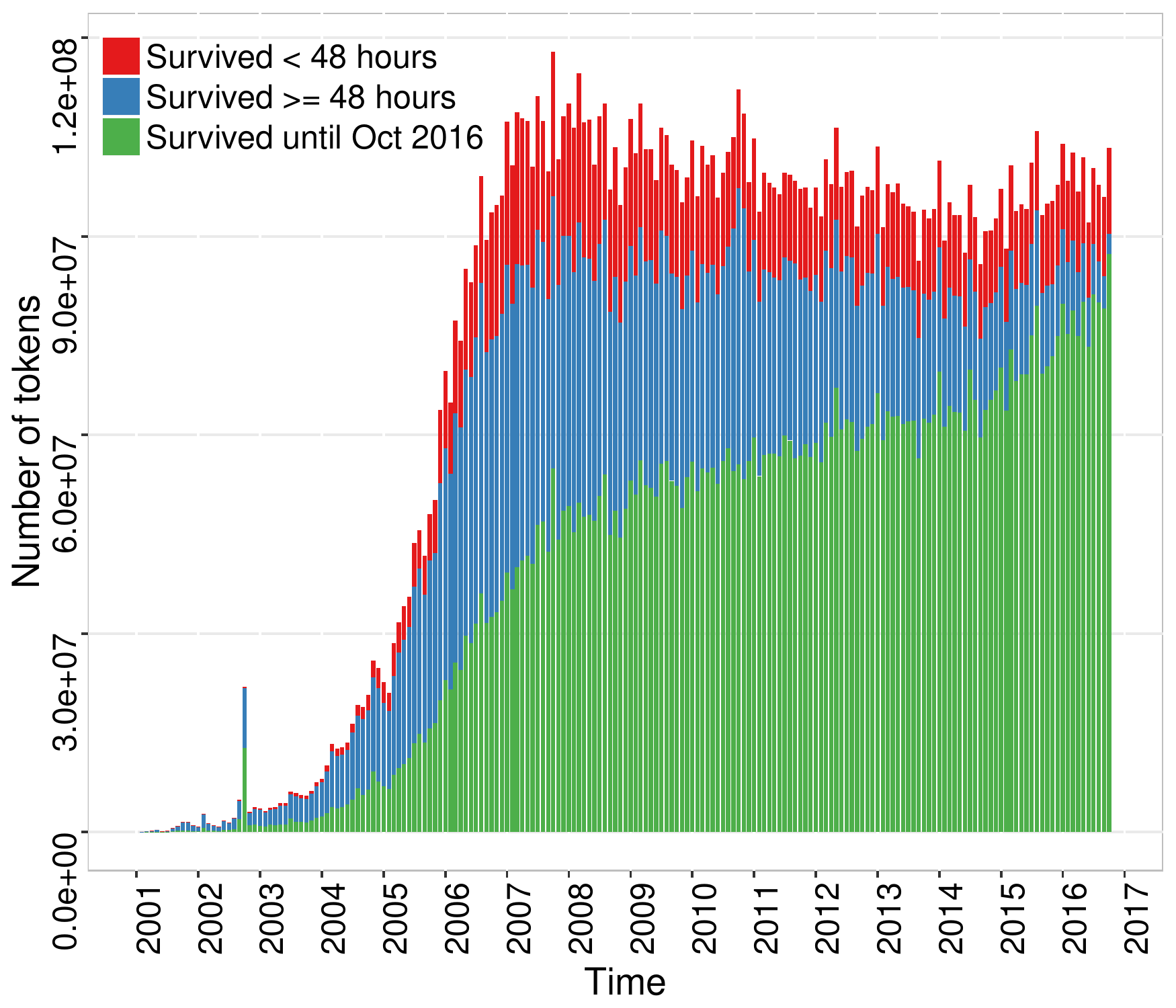}\\
\label{fig:sur_all}
}
\hfill
\hspace{2mm}
\subfloat[Tokens survived 48h and until Oct.'16 (cf. Fig. \ref{fig:sur_all}) in relation to newly added tokens, and the same ratio for the 48h-surviver-tokens per month from a top-1000 article sample.]{
\centering
\includegraphics[width=0.29\linewidth]{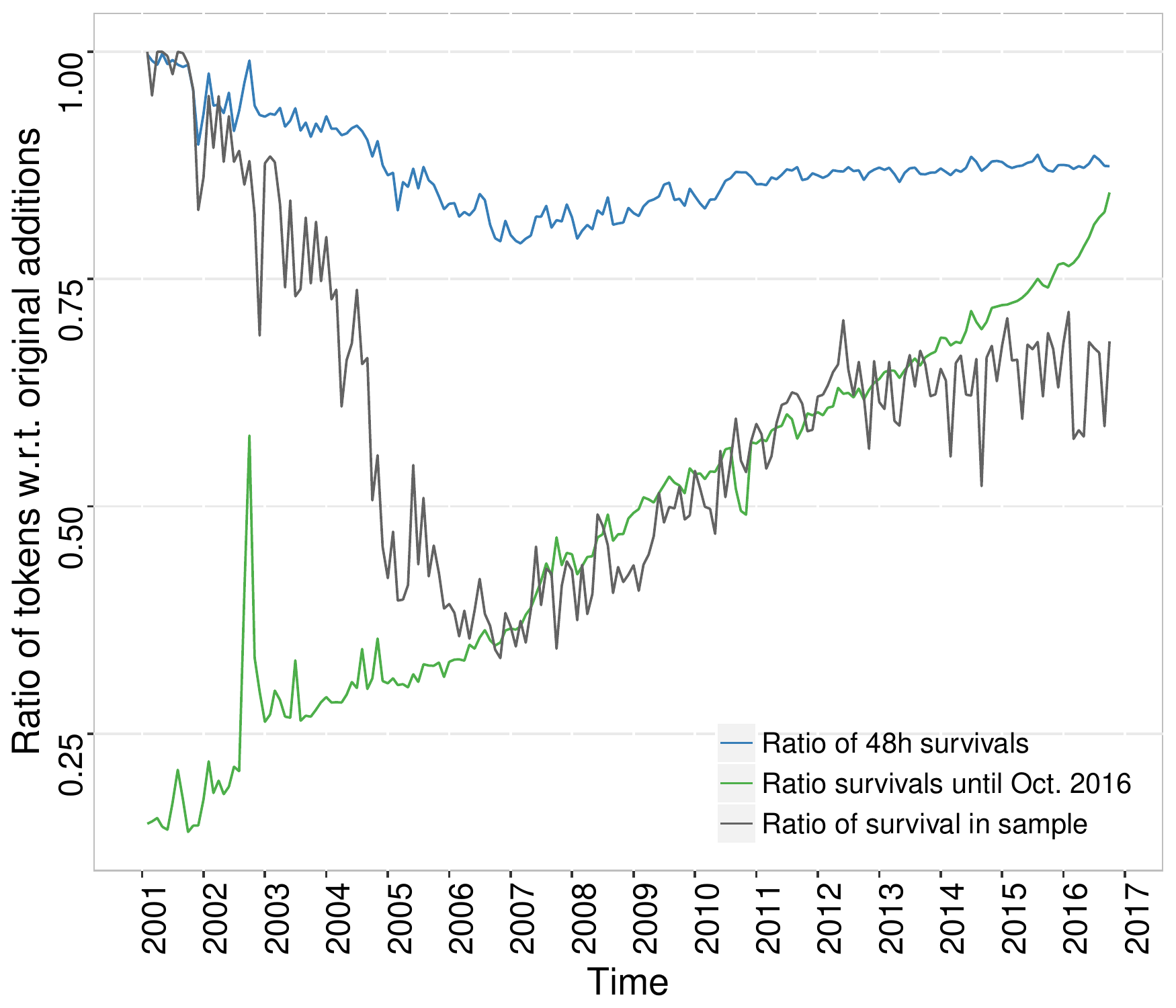}
\hspace{2mm}
\label{fig:sur_ratio}
}
\hfill
\hspace{2mm}
\subfloat[Tokens that survived 48h (as in Fig. \ref{fig:sur_all}) divided by the user group that added them: registered users, unregistered users or bots.]{
\includegraphics[width=0.29\linewidth]{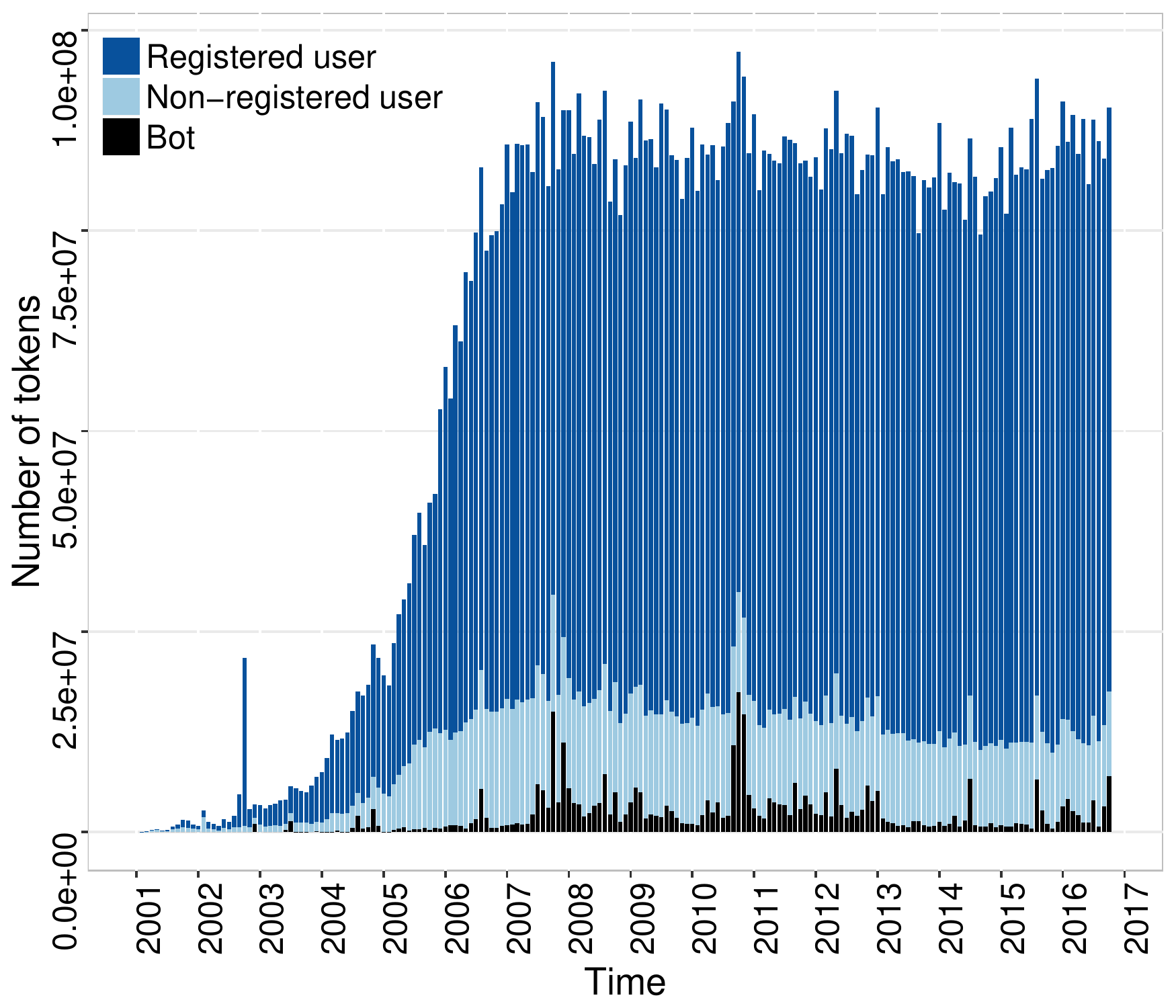}
\label{fig:sur_48h}
}
\caption{Survival of tokens over the whole history of the English Wikipedia.}
\label{fig:survival}
\vspace{-2mm}
\end{figure}

While looking at identity reverts is a viable and appropriate representation of interactions occurring between editors in these cases, 
Brandes et al. \cite{brandes2} make the argument for extracting \textbf{interactions based on all token-level changes} between editors (and thereby, revisions) when they state that identity reverts do ``not consider who deletes how much of whose edits" and which exact parts are reinstated later (cf. Section \ref{sec:datacreation}).
They construct a network of editors by taking into consideration if an editor (a) deleted tokens by another editor or (b) undid the deletions of another editor (negative interactions), which (c) results also in restoring the content of some third editor (positive interaction). 
Lerner and Lomi \cite{lerner2016dominance} extend this method to infer detailed disagreements and restorations between editors to study emerging hierarchies and we have earlier implemented an interactive visualization based on the technique proposed by Brandes et al. \cite{brandes2} and comparable data \cite{whovis,flock2015towards}. 
Maniu et al. \cite{maniu} infer a  very similar signed editor network, including deletion, revert and restore actions and tracking authorship of words, using text difference computation and edit comments. \textit{Contropedia} \cite{contropedia2} also employs interaction network visualization based on token changes.

This range of studies exemplifies how fruitful extracting editor interactions from revision data is for research. While not needed in every research setting, the minute interactions as used by \cite{brandes2} and others promise however to enable more precise explorations of these exchanges between editors, which can benefit future research. 
Such computations can be easily carried out on top of the TokTrack data, as we show in Section \ref{sec:usecases}, by extracting which tokens originating in which revision were removed or reinserted in which following revision. This, again, can be done more efficiently than with text-comparison algorithms, allowing to scale up the existing token-change-based approaches -- seeing that the ones we presented here only deal with up to a couple of hundred articles or single use cases, which we surmise partly to be due to the expense of calculation.











\section{Use Cases and Analyses}
\label{sec:usecases}

In this section we demonstrate how metrics related to the research strands in Section \ref{sec:relwork} can be extracted from our dataset and present the results of several analyses run on the full English Wikipedia content from the end of October 2016.\footnote{Data from the day of Nov. 1st, 2016 that was contained in the XML dumps is left out in the remainder.}   

 We will additionally publish the auxiliary datasets we created for each of the following analyses alongside the main dataset. These contain computed token survival information, conflict scores and revert data, to further facilitate the exploration of the TokTrek corpus. 

\subsection{Token Survival and Authorship}

The survival of tokens over time can give important insights about the resistance their introduction has faced, which might be due to different reasons, for instance their quality or features of the contributing editor \cite{halfaker_jury_2009}. We therefore first analyzed, based on all 13,545,349,787 
token change histories (i) the total number of tokens originally added in each month for the whole dataset (visible as the total height of the stacked bar charts in Figure \ref{fig:sur_all}). 
Also, as added tokens are often deleted again, we measured (ii) what number of tokens survived at least 48 hours as a gauge for their ``fitness'', which is represented by the sum of the blue and green bars in Figure \ref{fig:sur_all},\footnote{This cut-off is fitting as after 48h (or 5 revisions) the probability for deletion has reached very low levels, as has been argued  by Halfaker 
in a Wikimedia Research Showcase which inspired this particular analysis:  {\url{https://upload.wikimedia.org/wikipedia/commons/1/13/Anon_productivity_and_productive_efficiency_in_English_Wikipedia_(Showcase,_Jan._2016).pdf}}
-- see also: \url{https://meta.wikimedia.org/w/index.php?title=Research:Measuring_edit_productivity&oldid=16388960}
} as well as which tokens added in each month could still be read in Wikipedia at the end of October 2016, represented by the green bars in Figure \ref{fig:sur_all}.

\subsubsection{Analysis}
In Figure \ref{fig:sur_all}, we see that the rapid growth in added tokens leveled off around the beginning of 2007, and transformed into a slight decline before recovering towards the middle of 2014.  
As visible in Figure \ref{fig:sur_ratio}, the ratio of newly added content that was good or uncontentious enough to survive 48 hours exhibits a (mostly) continuous decrease from 2001 until 2007, coinciding with the change in total added content, then stabilizes and even begins to slightly climb again until recently. 
We compared these results to the ratio solely computed on a subset containing the 1000 articles with the most revisions. As seen in Figure \ref{fig:sur_ratio}, the more popular articles were subject to an even deeper drop of the likelihood of content to survive the first 48 hours, somewhat following the pattern for all articles, but not recovering to the same level.
 The survival rate of content from any month still present in Oct. 2016 is, naturally, near-linearly increasing for later months, as there is a higher chance for earlier tokens to be outdated and removed at one point, even if they survived the initial 48 hours. We see however a surprising spike in Oct. 2002 (also in absolute additions) and a slight drop between Oct. 2004 and July 2007 in survived additions that begs further examination in future work. 
 
Lastly, we split up the tokens that survived 48h in each month by the user group that originally added them: registered users, unregistered users or bots, as shown in Figure \ref{fig:sur_48h}. While it seems that the addition of persisting tokens of unregistered editors has become comparably stable since 2006, it has not been keeping up by far with the enormous increase by registered editors, which make up for over 80\% of all added surviving content for most months since 2007.
In fact, a small group of registered users generates the vast majority of sustained content (not depicted).
Bots showed an increased presence from mid-2007 until 2013, when, presumably by the migration of inter-language links to Wikidata, the demand for bot-created content dropped. 
 
Many questions remain open whose study we hope to enable through our dataset -- such as the longevity of  deletions, the role of bots in deleting content apart from adding it, or the survival of certain strings in specific article categories.


\subsection{Conflict}
To gain some first answers as to which parts of the content are conflicted, we present two metrics that can be straightforwardly computed from the TokTrack dataset. 
As the basis we used all $7,746,908,047$ 
tokens present at the end of October 2016 to get an overview of the  recently present content that has been controversial, excluding older vanished content. The following metrics were computed:
\begin{itemize}
	\item \textbf{Conflict Basic ($cB$)} sums up all deletion (``out") and reinsertion (``in") actions targeting a token over the whole revision history of an article.  We do not count the first deletion of a token, to avoid recording corrections 
that do not trigger a response
, similar to the mutual identity reverts proposed by \cite{sumi2011edit}, and  we also exclude undo-actions of editors on their own actions. E.g., token T5 from our toy example in Figure \ref{fig:trackxample} would accrue $cB=2$, as after the initial deletion, it was reinserted once and deleted again. If we 
suppose 
that R3 and R4 were submitted by the same editor, the last deletion would not be counted, resulting in $cB=1$. 
\item \textbf{Conflict Time-Aware ($cT$)} 
is almost equivalent to $cB$: instead of just counting up $1$ per undo action, it weights rapid undo actions higher by assigning them $\frac{1}{t_w}$.  The weight $t_w$ is computed as the logarithm to the base $3600$ of the absolute time $t$ in seconds that has passed since the last action on the token was performed.\footnote{In practice, we used $t+2 sec$, since two revisions can be recorded with the same timestamp. This guarantees that $t_w > 0$.} Thus, up to 1 hour time difference, undo actions are weighted considerably higher than the original time in seconds and their weight decays fast, while after 1 hour they are weighted lower, but the weight decay is decreased.  Supposing for T5 in Figure \ref{fig:trackxample} the time between R2 and R3 is $20$ seconds
, we would hence get $cT=2.73$ for T5 (again assuming R3 and R4 are by the same editor).

\end{itemize} 

The purpose of these simple metrics is to demonstrate how our dataset can serve to give an overview of conflict for the whole English Wikipedia -- the first ever  based on all individual token changes, as far as we can tell. 
Future work will undoubtedly conceive more intricate token-based measures or adopt such as proposed by \cite{bykaufine}. One could also for instance imagine a weighting of undo actions by the experience of the editor in the article as per \cite{sumi2011edit}, to more effectively down-weigh vandalism, or the editor's  overall proneness to conflict \cite{vuong_ranking_2008}.



\begin{table}[t!]
\centering
\caption{The top 15 most conflicted articles in the English Wikipedia end of Oct. 2016, as per the sum of $cB$ (left) and $cT$ (right) assigned to their currently present tokens.}
\includegraphics[width=0.79\linewidth]{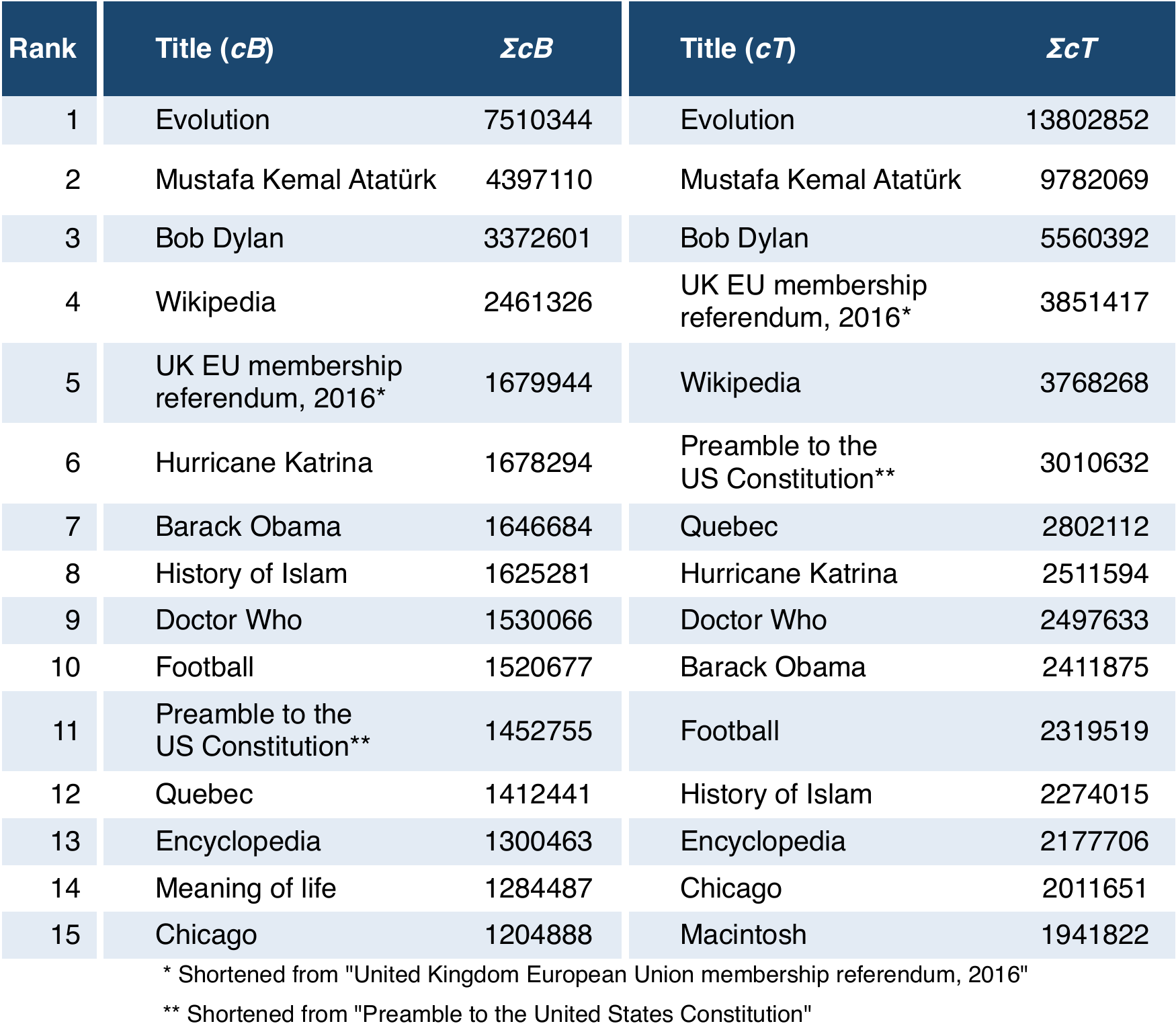}
\label{tab:con_articles}
\vspace{-3mm}
\end{table}


\subsubsection{Analysis}
As reporting the conflict scores for individual token instances would be too detailed, we present two aggregations: most conflicted articles and most conflicted string values for the whole English Wikipedia. 

For Table \ref{tab:con_articles}, we (i) separately summed up the $cB$ and $cT$ scores of the tokens present per article and (ii) ranked the articles by the $\sum cB$ and $\sum cT$ of all their tokens. We show the top 15 of both ranked lists. We see that while the rankings differ, most articles are shared by both top rankings. For the complete ($>5$ Mio.) article lists, the correlation is indeed very high (Spearman's $\rho=0.99$, Pearson's $\rho=0.98$), 
indicating that the time-weighting does not have a very strong impact, although it is certainly notable.
The topics feature some ``evergreens", as also included in the top 100 most controversial articles of \cite{yasseri2014most},\footnote{\label{fn:warlist}Retrieved from \url{http://wwm.phy.bme.hu/Top100/top100_en_wiki.txt}, as referenced in \cite{yasseri2014most}, who did not limit their analysis to current content or reverts.} such as  {\sf \footnotesize Evolution}, {\sf \footnotesize Quebec}, and {\sf \footnotesize Wikipedia}, but naturally have a strong recency bias, as only currently still present tokens were used.\footnote{For a detailed study of temporal developments of controversies, the  extraction of timestamp-based sequences of deletions and reinsertion from TokTrack should be used.} We can for example see the article about the {\sf \footnotesize United Kingdom European Union membership referendum, 2016} (commonly known as ``Brexit") on top in both lists, as well as the article about musician-turned-nobel-prize-2016-winner {\sf \footnotesize Bob Dylan}. The article of 2016 U.S. presidential candidate {\sf \footnotesize Hillary Clinton} ranks at 16 ($cB$) and 22 ($cT$), with {\sf \footnotesize Donald Trump} trailing at rank 33 and 31, respectively.

Next, we summed up the conflict values $cB$ and $cT$ of all unique strings 
from the token instances that carry those string values.  Naturally, very common strings such as ``to" accrue very high overall sums; 
we hence normalized the values by dividing by the overall frequency $n$ of tokens with the same string value, yielding $\frac{cB}{n}=cB_n$ and $\frac{cT}{n}=cT_n$. 
Table \ref{tab:con_strings_all} shows the top 15 most conflicted string values over all articles by both $cB_n$ and $cT_n$ that appear at least $n=1000$ times.\footnote{$n \geq 1000$ to exclude many very rare tokens that are hard to interpret here and get ranked high in the normalization process.} 
Again 
the correlation between both full lists of strings is very high (Spearman's $\rho=0.88$, Pearson's $\rho=0.96$).
\begin{table}[t!]

\caption{Top most conflicted string values and an example of most conflicted strings in an article.}
\subfloat[Top $15$ most conflicted strings with $>1000$ appearances in the complete English Wikipedia end of Oct. 2016, ranked by $cB$ (left) and $cT$ (right). The strings ``dumbledore'' and ``voldemort'' are explored more in detail (cf. text).]{
\includegraphics[width=0.49\linewidth]{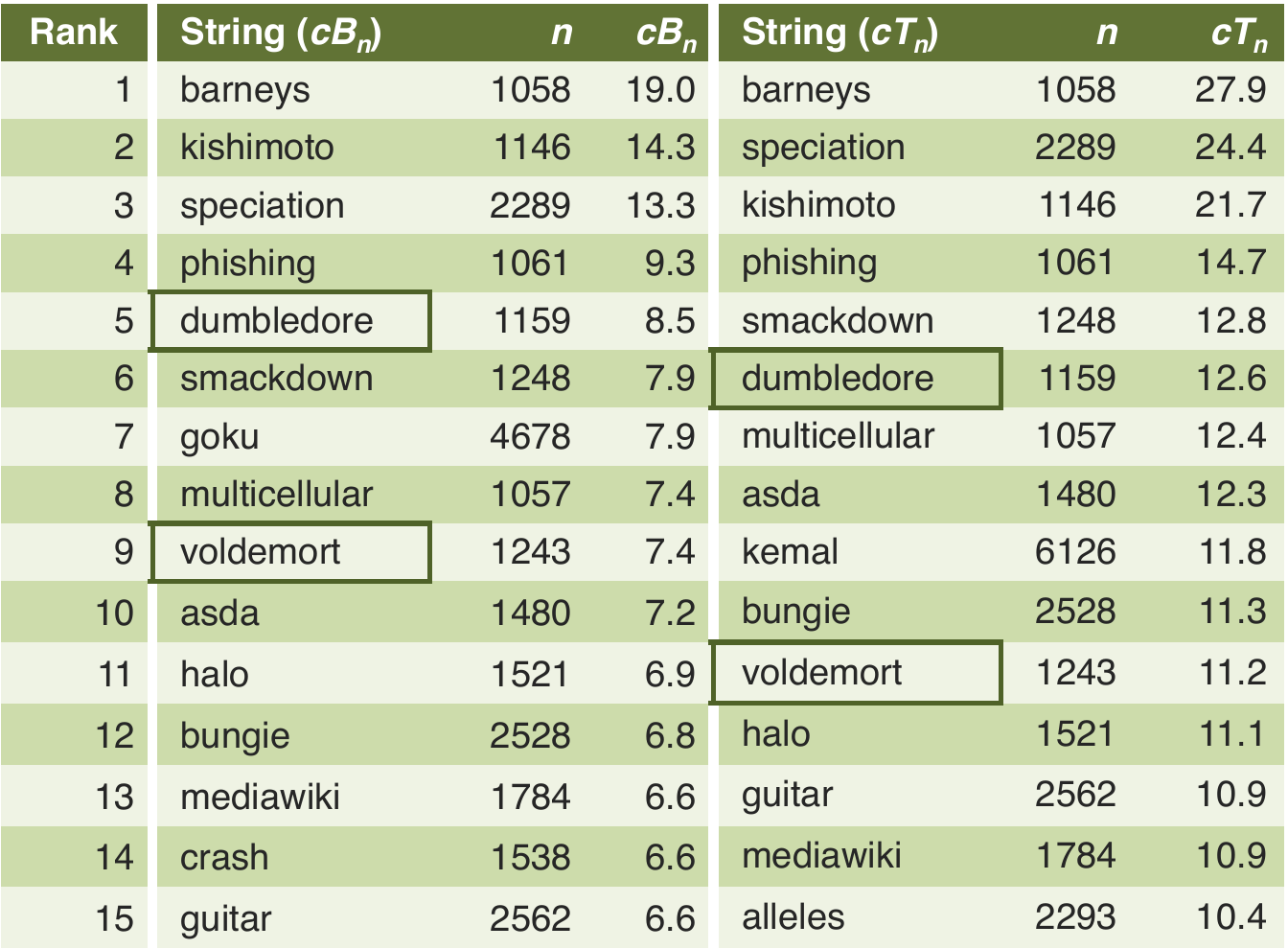}
\label{tab:con_strings_all}
}
\hfill
\subfloat[Most conflicted strings in {\sf \scriptsize United Kingdom European Union membership referendum, 2016}, ranked by $cT_n$. The right-hand ranking ignores strings with $na<10$ in the article.  \newline]{\includegraphics[width=0.45\linewidth]{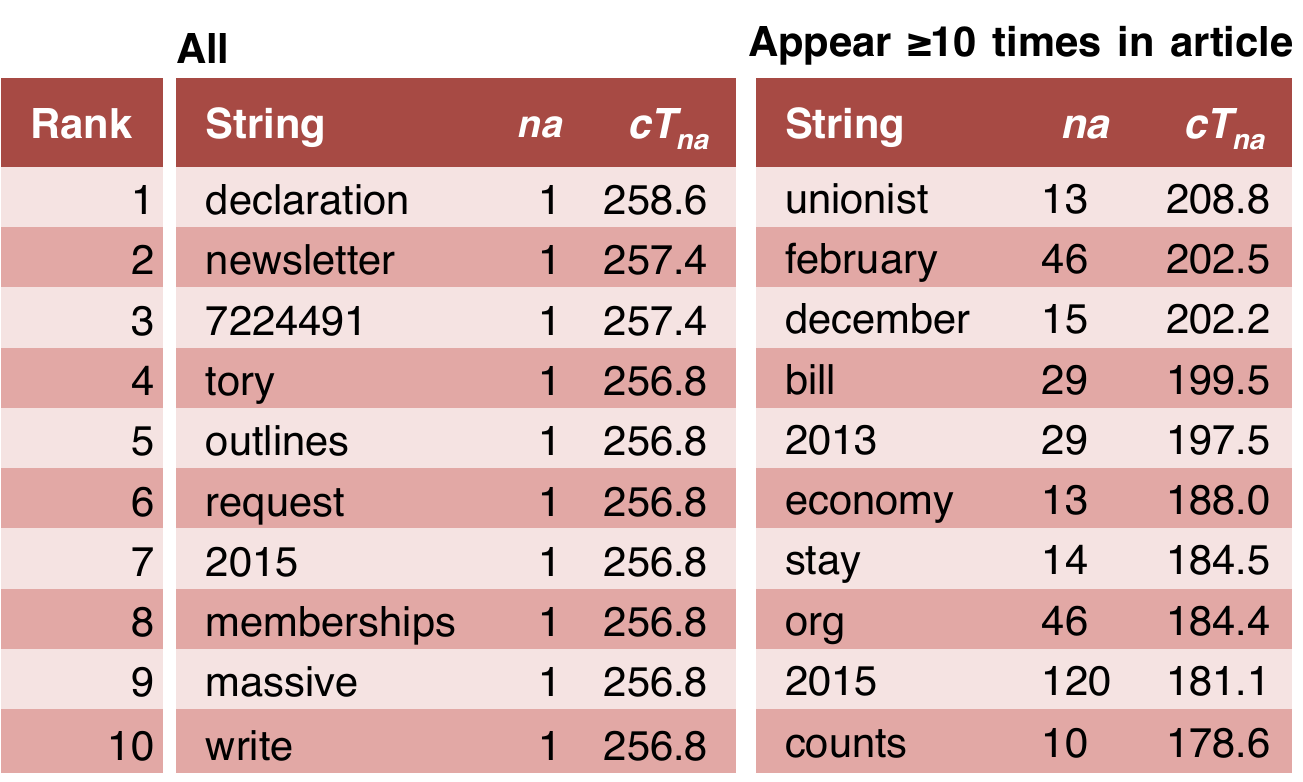}
\label{tab:con_strings_brexit}
}
\label{tab:con_strings}
\vspace{-2mm}
\end{table}
As our dataset also allows to explore in which concrete contexts (token instances) a string has been most controversial, we investigate the strings  ``dumbledore'' and ``voldemort'' (names that rather uniquely identify two literary characters; marked in Table \ref{tab:con_strings_all}). The top 30 articles in which they are most controversial all span a range of topics from the ``Harry Potter" literary universe, (from {\sf \footnotesize Rubeus Hagrid} to {\sf \footnotesize Magical objects in Harry Potter}), pointing to conflicts around the two names. 
Most revealing, however, is that while most of these articles rank somewhat high on the conflicted article lists (average rank of around $6500$), the two terms in comparison show much higher conflict in the string ranking.  That we could single out these two disputed terms demonstrates how it is possible to spot very localized controversies with the token-based approach, which might not be mirrored in the overall controversiality of single articles, but can evolve ``horizontally'' over several pages.\footnote{Surely, due to our na\"{i}ve normalization by global frequency, very common terms that were disputed locally can lose visibility here. A more elaborate approach should take this into account.}
 
 Lastly, we take a closer look at the ``Brexit" article as identified in Table \ref{tab:con_articles}. To this end we computed the local sum of $cT$ of all tokens in this specific article and normalized it not by $n$, but the frequency $na$ of the strings only in this article, giving us $cT_{na}$ for each string.
   Table \ref{tab:con_strings_brexit} ranks the strings in the article by $cT_{na}$ for the top 10. For less common strings we see some conflicts around ``declaration", the name of the Tory party, and a number that is part of a reference URL. More frequently used strings under contention are ``unionist" as part of two parties' names, as well as ``bill'' and several date-related strings. We thus get a better understanding of the controversies in the article. The next step would be to set these string instances into the context of other controversies in the article or their Wikipedia-wide conflict scores -- data that is readily available from TokTrack.


\subsection{Interactions Between Users}

To show another use case for the dataset, we studied how many ``undo actions'' similar to the editor interactions defined in \cite{brandes2} can be extracted. 
We use the term ``edit action" to refer to three different types: 

$Add$: adding a completely new token, 

$Del$: deleting a single token, 

$Re$: performing the reinsertion of a single token.

\noindent The last two types, $Del$ and $Re$, are always considered ``undo actions'', as they undo either a previous addition of a token ($Del$) or undo the deletion of a token ($Re$).\footnote{$Re$ is at the same time always a ``redo action'' toward the revision (and in extension, editor) that added the token the last time. 
Yet, we will not focus on this aspect here.}
One edit (creating a new revision) can hence possibly contain multiple edit actions, e.g., removing $n$ tokens ($n \times Del$ actions) and in their place adding $m$ completely new ones ($m \times Add$), which would amount to $n+m$ edit actions (i.e., we also do not define an explicit ``replace" action). 
 
 Illustrated on our toy example (cf. the arrows in the table of Figure \ref{fig:trackxample}), revision R4, e.g., undoes 4 actions by R1 (of 6 actions R1 did originally) and 4 actions by revision R3 (of also 6 originally by R3).
 This means R4 carries out a 4/6 \textit{partial revert} each on two target revisions (i.e., 8 total actions). If \textit{one} individual revision undoes \textit{all} edit actions by a preceding revision we call this a \textit{full revert} in the remainder.


%

\begin{table}[t!]
\caption{Amounts and types of reverts extracted from all articles ($451,350,901$ revisions): 
Showing \textit{unique} reverting and reverted revisions.  Top rows: revisions that reverted a revision by another editor. Bottom rows: self-reverts. 
}
\centering
\begin{tabular}{lrr}
\toprule
{\bf Revert} & {\bf Reverting} & {\bf Reverted} \\
{\bf  type} & {\bf  revisions} & {\bf revisions} \\
\midrule
{\bf Non-self full} & $59,895,674$ & $88,081,312$\\
{\bf Non-self partial} & $184,344,591$ & $170,356,394$\\
\midrule
{\bf Self full} & $7,461,983$ &  $8,137,440$\\
{\bf Self partial} & $59,723,545$ & $49,892,931$\\
\bottomrule
\end{tabular}
\label{tab:reverts}
\vspace{-2mm}
\end{table}

\subsubsection{Analysis}
We ran the computation to extract these undo actions and reverts over all 
 articles and revisions, checking the changes to all  $>13$ Billion tokens in TokTrack.

Table \ref{tab:reverts} shows the total number of revisions that have undone all actions of another editor's revision completely (``non-self full'') or only partially (``non-self partial''); as well as the full and partial reverts that editors have carried out on their own revisions (``self''). Note that reverting revisions can always target multiple revisions (full or partial); and while the target of a full revert is only targeted once -- then its actions are undone -- partially reverted revisions can be undone by several revisions. The latter explains the lower counts of partially reverted revisions in Table \ref{tab:reverts}: content added by one revision can over (a long) time be corroded by many small changes. In this light, since we did not introduce a revision limit or time limit for scanning backwards in an articles/tokens history, ``revert'' cannot per se be equated with antagonism here, as these numbers include the complete spectrum from minor corrections to full-on opinion clashes and vandal fighting.

In total, $61.51\%$ of all edits included some kind of removal or reinsertion of content (i.e., $38.49\%$ revisions purely added content), and in $14.62\%$ 
of the revisions editors correct their own edits. 
$14.84\%$  
of all revisions fully undid another revision and $50.65\%$ did so partially. Per our definition, revisions that perform full reverts can also carry out other actions simultaneously, leading to unique new revision content. That is likely the reason that the identity revert method which we ran over the same articles and revisions in comparison marked $5.97\%$ 
fewer of the revisions ($8.87\%$; $40,033,526$) to be fully reverting other revisions.\footnote{Based on the SHA hash values for each revision's content. The revisions between two identical ones are considered fully reverted by the second identical revision. Partial reverts are not detectable.} These results seem to confirm that most \textit{fully reverting} revisions are identifiable with the identity revert method as prior research has assumed. However, when looking at  the distinct pairs of reverting-reverted revisions the identity method detects, only $73.45\%$ of those are found as full reverts in TokTrack as well and $12.98\%$ are identified as partial reverts (with the rest not judged to be reverts at all). This hints at the strong possibility of misidentifying \textit{type} as well as the \textit{targets} of reverts when using the identity revert method, as we have argued previously \cite{floeck2012revisiting}. Especially when constructing editor interactions   from such data, this has to be considered.




\begin{figure}[t!]

\subfloat[Number of revisions that undo a specific ratio of the edit actions of another revision.]{
\includegraphics[width=0.48\linewidth]{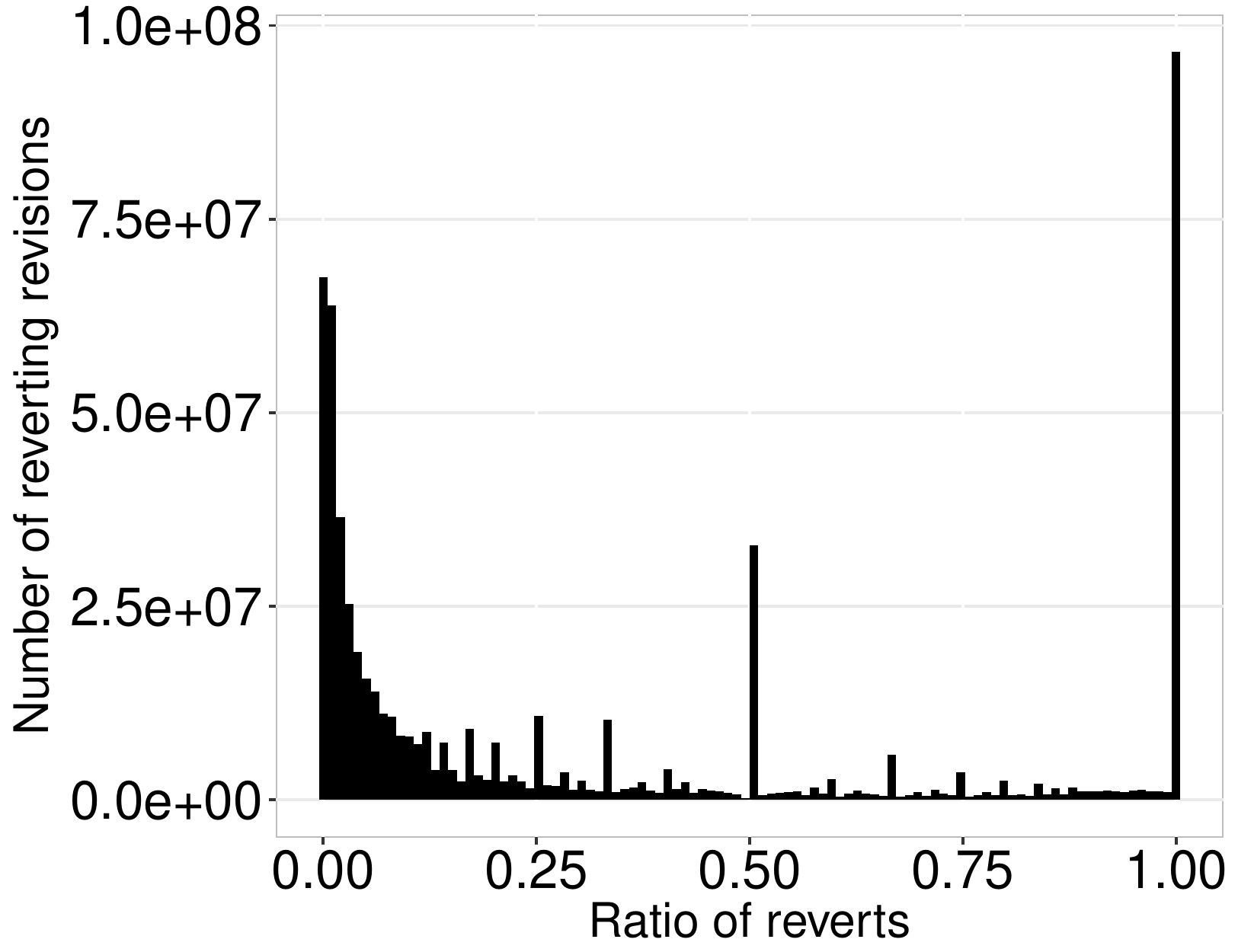}
\label{fig:rev_ratio}
}
\hfill
\subfloat[Number of reverting revisions that undo a specific absolute number of edit actions.]{
\includegraphics[width=0.48\linewidth]{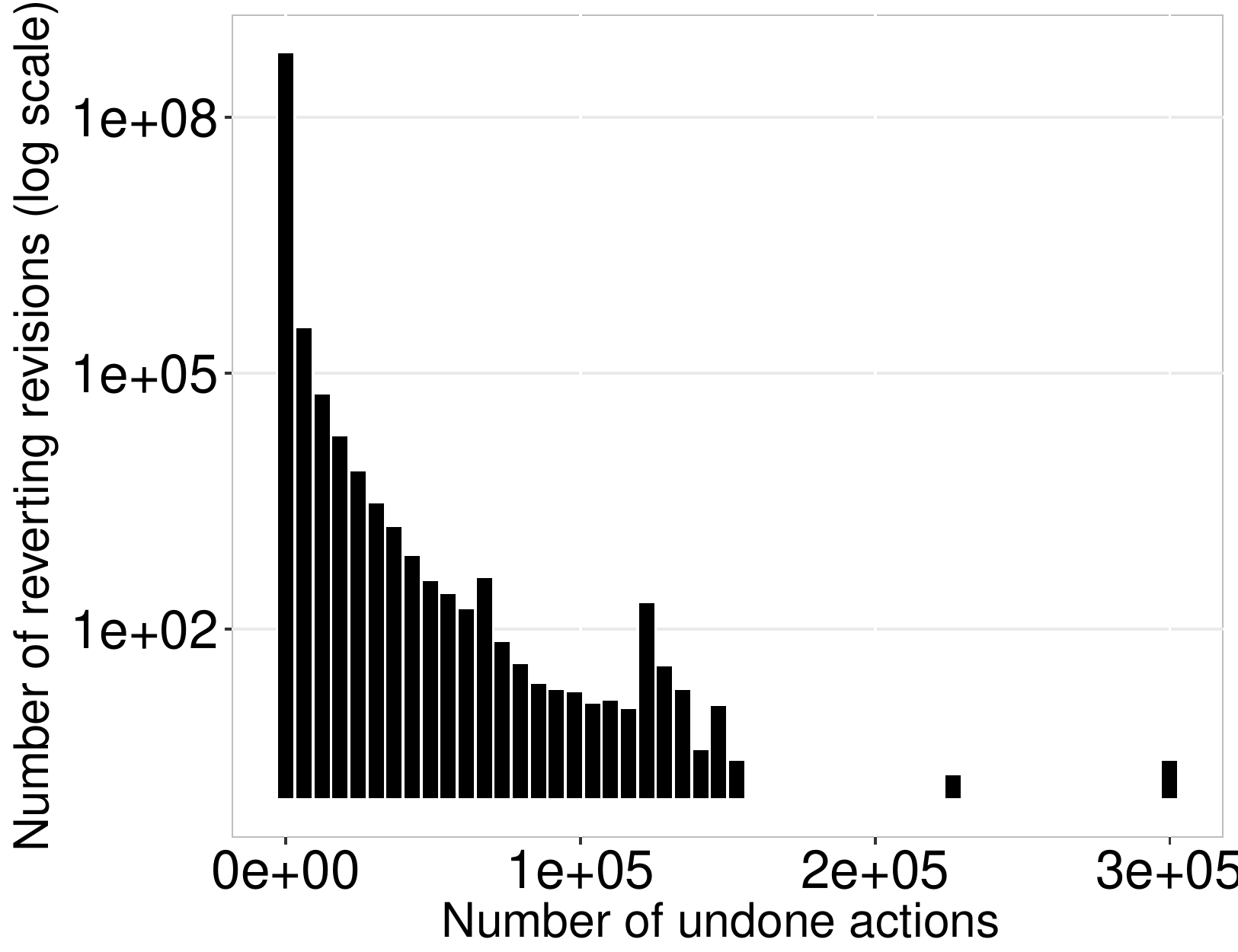}
\label{fig:rev_absolute}
}
\caption{Numbers of revisions that undo a certain amount or a ratio of edit actions}
\vspace{-3mm}
\end{figure}

As partial reverts naturally make up the largest part  of the revisions undoing others 
and have not been explored much in related studies, we take a closer look at them. Figure \ref{fig:rev_ratio}  shows which ratios of reverts are most common. Apart from full reverts  making up a large share, in partial reverts, smaller-ratio corrections are generally much more common than larger ones. The spike around the $50\%$ ratio seems odd at first. Yet, after we investigated a large number of these cases by hand, we saw that
these changes incrementally replace (improve?) content that was already a replacement for even older content. E.g., the addition of one token ``B",  is undone, to put in ``C", but the deletion of the older token ``A" -- which ``B" has replaced -- is not undone. 
These incremental corrections seem to be very common and often only encompass very few undo actions. 
Further, such partial reverts that \textit{might} be considered as mostly disagreeing because they undo more than $50\%$ (and $<100\%$) of their target revision's actions make up only $9.46\%$ of all reverting revisions.
From Figure \ref{fig:rev_absolute} we can moreover glean that the absolute number of undone actions per reverting revision is for the largest part very low. 
As a last insight we considered a subset of the 1000 articles with the most revisions in Wikipedia.  For this sample, the proportion of (all) full reverts to all revisions rises to 25.28\% (from 14.84\%) 
compared to the full article set, 
and partial reverts with over 50\% undo rate rise to 18.52\% -- possibly indicating a higher disagreement level, which might be a result of higher popularity of these articles.

In summary, these results shed some light on partial reverts and corrections, self-corrections, and the differences between revert detection methods, insights so far not reported for the whole English Wikipedia, to our knowledge.



\section{Conclusions}
With the dataset presented here, we hope to remove some of the larger barriers when it comes to computing content-based, fine-grained metrics on top of Wikipedia's revisioned content that enable insights into its collaboration dynamics and the way its content evolves under the influence of these social patterns. We showed how the token-level data can help to understand existing findings better as well as how it can open new avenues for research on Wikipedia content. We aim to extend our work to other language editions and projects in the foreseeable future.

\bibliographystyle{abbrv}
\fontsize{9.0pt}{10.0pt}
\bibliography{sample}

\end{document}